# A Behavior-based Approach for Multi-agent Q-learning for Autonomous Exploration

Dip Narayan Ray [#1], Somajyoti Majumder [#2], Sumit Mukhopadhyay [*3]

[#] *Surface Robotics Lab, Central Mechanical Engineering Research Institute (CSIR)*
*Durgapur, W.B. 713209, India*

[1] dnray@cmeri.res.in
[2] sjm@cmeri.res.in

[*] *Department of Mechanical Engineering, National Institute of Technology*
*Durgapur, W.B. 713209, India*

[3] msumitnit@yahoo.co.in

*Abstract* — **The use of mobile robots is being popular over the world mainly for autonomous explorations in hazardous/ toxic or unknown environments. This exploration will be more effective and efficient if the explorations in unknown environment can be aided with the learning from past experiences. Currently reinforcement learning is getting more acceptances for implementing learning in robots from the system-environment interactions. This learning can be implemented using the concept of both single-agent and multiagent. This paper describes such a multiagent approach for implementing a type of reinforcement learning using a priority based behaviour-based architecture. This proposed methodology has been successfully tested in both indoor and outdoor environments.**

Keywords: Multiagent, Reinforcement learning, Q-learning, Behavior-based robotics, Autonomous exploration

## I. INTRODUCTION

Recently the field of robotics, especially the mobile robotics has been identified as on of the most important areas of research due to its huge potential for autonomous explorations in different hazardous or toxic and unapproachable domains. These exploration domains extend from underwater exploration to factory automation, polar to planetary exploration, landmine detection to unknown environment mapping. But for such explorations, the use of a mobile robot with classical control is possible if and only if the programmer or user has the prior knowledge about the environment. It is completely impossible to develop a mobile robot for explorations without knowing the environment beforehand. For such cases, the concept of learning from past experiences may provide a better strategy for explorations. The system will learn constantly from the interactions with the environment and modify the strategy of exploration accordingly. The most suitable learning in this direction is the Reinforcement learning, especially the Q-learning which uses delayed rewards [1]. The current research work proposes a new approach of autonomous exploration using multiagent Q-learning using behaviour-based robotics. This paper is organised as follows: after this introduction, related works and a few insights have been described. Then there are proposed methodology and experimental results and discussions followed by a conclusion.

## II. PREVIOUS WORKS

The field of reinforcement learning [1] has been started only few years ago. Reinforcement learning is being used for various multiagent systems to solve problems with numbers of robots/ systems. Lots of works have been done mainly theoretically and using simulation. The following paragraphs describe the work done in the field of MARL.

Reference [2] uses the traditional Q-learning algorithm for multiagent system. The main interest is focused on the complex behavior of Q-learning with $\mathcal{E}$-greedy exploration in Prison-Dilemma-like games and the algorithm is able to achieve higher-than-Nash outcomes in an undiscovered chaotic system. Panait and Luke have wrote a state-of-the-art paper [3] discussing the co-operative multiagent learning in broad survey. But it clearly mentions that Reinforcement Learning methods have only theoretical proof of convergence and such convergence assumption do not hold for some real-world applications including many multi-agent system problems. The application of multi – agent for Q-learning was in [4] with the use of "Q Updates with Immediate Counterfactual Rewards-learning (QUICR – learning) algorithm which is designed to improve both the convergence properties and performance of Q-learning. This proposed modification tries to solve the existing credit assignment problem for a multiagent system. Reference [5] reviews the main benefits and challenges of multiagent Reinforcement Learning (MARL) as well as the different viewpoints on defining the MARL learning goal. There is also discussion of MARL algorithms for fully cooperative, fully competitive and mixed tasks. Issues on how autonomous agents learn to solve dynamic tasks online have been discussed. Behaviors of several MARL algorithms have been studied in simulation environments. Littman [6] described an approach to reinforcement learning in multiagent general-sum games in which learner is told to treat each other agent as either a friend or foe. This algorithm also provides strong convergence than Nash-equilibrium-based learning rule. Case-Based Heuristically Accelerated Multiagent Reinforcement Learning (CB-HAMRL) has been proposed by Bianchi & de Màntaras [7]. This algorithm is based upon an emerging technique, Heuristic Accelerated Reinforcement Learning, in which RL methods are accelerated by making use of heuristic information. Empirical evaluation has been conducted in a simulator for the Littman's robot soccer domain. Fujii et all [8] have used multilayered reinforcement learning scheme to select the appropriate collision avoidance behaviors so as to reduce the computational power and memory capacity. This helps to move numbers of robots using LOCISS (Locally Communicable Infrared Sensory System) safely in an environment full with large numbers of static obstacles. They have also used the algorithm on a real robot. This paper [9] aims to build a team of agents on long term basis where the decision making will be done using reinforcement learning. It uses robotic soccer as a multiagent Markov Decision Process and the analysis is made in own-developed 'Karlsruhe Brainstormers' simulator. Discussions on how optimality of behaviors of agents can be defined and the difficulties associated with developing algorithms to reach such optimality have been discussed. Kim and Vadakkepat [10] have used three micro-robot soccer teams in real field for analyzing the multiagent systems from robot soccer perspective. They have also reviewed the multiagent system and the learning issues in multiagent systems from robotic soccer perspective. The action selection [11] for the cooperation and coordination among agents is an important issue for multiagent systems. In dynamic and complex environments, the modular Q-learning is proposed for selecting the right action. The modular Q-learning consists of two different parts: different learning modules and the mediator module. The mediator module selects the right action based on the Q-values obtained from different learning modules. Littman [12] has described a Q-learning like algorithm for finding optimal policies for a two agent systems with diametrically opposite goals. It basically uses a reinforcement learning approach to solve two-player-zero-sum games in which the max operator in the update step of a standard Q-learning algorithm is replaced by a minimax operator that can be evaluated by solving a linear program. Reference [13] describes a framework for using reinforcement learning on mobile robots. The main feature of the work is the use of example trajectories to bootstrap the value function approximation and splitting learning into two different phases. The first phase uses human or an example for the task and reinforcement learning system observes the states, actions and rewards by the robot. When the approximation of the value-function is done properly, the reinforcement learning in the second phase is in the control of the robot as in the standard Q-learning framework. Reference [14] is a review paper which focuses on the relation of multiagent reinforcement learning (MARL) with stochastic games. This paper has also discussed varieties of topics under MARL, ranging from proof of convergence of algorithms theoretically, develop algorithms of multiagent systems and simulate them, develop robots work to achieve certain tasks or develop social skills. The author has also hoped that combination of traditional RL solutions, applied MARL research, decision theory and game theory will open a new horizon of research.

It may be concluded from the above discussions that three types of works are reported in literature, for both single-agent and multiagent reinforcement/ Q-learning.

1) First type of papers [3, 5, 13, 14] is basically review type and discuss about the work done so far in the field. They do not propose any theory or describe any experiment.

2) Second type of papers [2, 4, 6, 7, 9, 11, 12] is theoretical based and the proposed methodologies/ algorithms or any modifications of existing algorithms have been established by simulation. Further more such types of papers can be categorized into (a) purely analytical [2, 4, 6, 11, 12] and (b) simulation based robotics [7, 9].

3) Third type of papers [8, 10, 13] are experimental based i.e. the papers discuss the use of real robots in indoor/simulated environments, although work is very limited.

Literature survey also reveals that till date no work has been carried out using robot for outdoor explorations using single agent/ multiagent reinforcement learning. The current work has tried to address this issue of autonomous outdoor exploration using multiagent Q-learning based on behavior-based robotics.

### III. A FEW INSIGHTS

This current work is related mainly with behavior-based robotics, reinforcement learning (especially Q-learning), multiagent system. The following paragraphs will provide a brief idea about the above topics in nutshell.

### A. Behaviour-based Robotics

The existing conventional/classical robotics has some control mechanism, which guides the end effectors to act accordingly, after it analyzes the inputs, obtained from various sensors and sends responses to those end effectors. The inputs from various sensors are used intermediately to symbolically represent the environment or action. On the basis of this environmental model planning is done and then only commands are sent to the end effecters. But if the end effecters are directly coupled to those sensors and there is an intelligent agent to control the system individually, then it will be able to take decision itself. So, this is one kind of intelligence, often looked for in robots. This behavior is often called 'Reactive' in nature like the closing of eyes due to intense light in human beings.

In behavior-based robotics four architectures are popular all over the world. They are Subsumption Architecture ([15], [16], [17], [18]), Action Selection Dynamics [19], Schema-based approach ([20], [21], [22], [23]) Process Description Language [24]. Out of which only the subsumption architecture has been used in the current research for implementation. Subsumption architecture is a layered behavior proposed by Brooks. These layers are associated with many simple behaviors. All these simple behaviors combine to form complex behaviors. The layers operate asynchronously.

### B. Reinforcement learning and Q-learning

Learning is needed for forming a map/ function between the state (sensor information) and action (actuator commands). Supervised learning approaches need the model of good behaviour by a teacher. Reinforcement learning is one of the widely used online learning methods in robotics. It is sometimes called learning with the critic that gives scalar reward or punishment based on behaviors [25]. The robot learns during an action and also acts/ responds during the learning, which is neglected in supervised learning methods. The Reinforcement learning method used in most of earlier works is single agent based Q-learning [26, 27]. In single agent based Q-learning algorithm the external world is modelled as Markov Decision Process with discrete states and action spaces. It is flexible in this sense because it can learn from actions which it/ programmer doesn't suggest. This ability is often called exploration – insensitivity. After each step an agent (single) observes the state vector $x_i$, chooses and applies an action $a_i$. The system passes in state $x_{i+1}$ and the agent receives a reinforcement $r(x_i, a_i)$. The goal of the learning is to find a policy which maximizes the sum of the future reinforcements. For a given policy $\pi$, it is noted that $a_i = \pi(x_i)$, the chosen action. The evaluation function $\pi$, noted $V^\pi$, is given by:

$$V^\pi(x) = E\{r_0 + \gamma r_1 + \gamma^2 r_2 + \dots \mid x_0 = x; \pi\}$$

$$= E\left\{\sum_{i=0}^{\infty} \gamma^i r_i \mid x_0 = x; \pi\right\}$$

(1)

The discounting factor $\gamma \in [0, 1]$ smaller than 1, ensures convergence of the sum. The optimal evaluation function V* (x), is defined as follows:

$$V^*(x) = \max_{a \in A_x} \left[ r(x, a) + \gamma \sum_y P_{xy}(a) V^*(y) \right]$$

(2)

$$= \max_{a \in A_x} Q^*(x, a)$$

Where: $A_x$ = set of possible actions in the state $x$; $P_{xy}$ = probability of passing from the state x to y by the action a; Q(x,y) represents the total reinforcement if the action a is selected in state x and if an optimal policy is chosen thereafter. It is called the quality function. Optimal policy can be found using dynamic programming if transition probabilities $P_{xy}(a)$ and the reinforcement law r(x, a) are known. Instead of using evaluation function, Watkins proposed to estimate the function Q* by the function:

(x, a) → Q(x, a) which is updated with each transition by:

$$Q(x_t, a_t) \leftarrow Q(x_t, a_t) + \beta \left[ r(x_t, a_t) + \gamma \max_{a \in U_{x_{t+1}}} Q(x_{t+1}, a) - Q(x_{t+1}, a_t) \right]$$

(3)

Here $\beta$ is a learning parameter which must tend towards 0 when t tends towards the infinite.

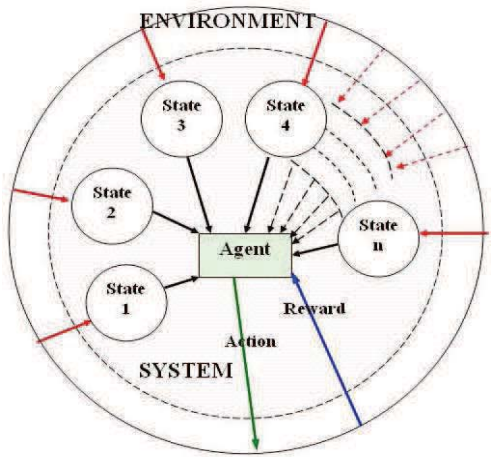

Fig. 1. System- Environment Interaction Model for Single agent Q-learning

### C. Multiagent Systems

Agents are the most discussed issue for present-day advanced robotics. However this concept of agents has evolved from multi-agent system (MAS) [29] which in turn is a part of Distributed Artificial Intelligence (DAI). The (software) agent is an entity which performs continuously and autonomously particular task(s) assigned to it in an environment (populated by other agents) for achieving the desired goal. Multi-agent system is composed of multiple interactive intelligent agents that collectively complete their own individual goals to achieve the overall goal. Multi-agent system (MAS) is very useful to accomplish a complex task easily and fast. In the most general case, agents will be acting on behalf of users with different goals and motivations. To successfully interact, they will require the ability to cooperate, coordinate, and negotiate with each other, much as people do.

### IV. THE PROPOSED METHODOLOGY

The proposed methodology uses the concept of gradual learning on a single agent and then a multiagent system. However the major work has been done on the multiagent system.

### A. Gradual learning

The term "learning" straight forward refers mainly to human learning. Many researchers have proposed different learning models to describe human learning. It has also been mentioned that human learning relies on the past experiences. For example, the learning of day -5 depends on the learning of day-4. One cannot learn the knowledge of day-5, without the knowledge of day-4 or day-2. The knowledge of a human being is refined gradually depending up on his/ her past experience, interaction with the environment, environmental conditions. Another important inference can be drawn from the theories about human learning is that learning is in terms what he/she already knows [30]. If one doesn't understand what does 'capital' or 'country' means, he/she cannot understand that "'A' is the capital of the country 'B'". So there is one popularly used method of learning which is very much gradual/ cumulative rather than sudden. For example a child starts learning the alphabets; if first day he/ she starts learning from 'A' and stops at 'D', next day he/ she will start from 'E', not again from 'A'. This saves time and increases the efficiency of learning.

### B. Human-like Gradual Multiagent Q-learning

This type of gradual learning can also be implemented for robot learning. In that case, the past experience of the robot can be referred for new or advanced learning. Q-learning, a well-known and most suitable RL for robots uses only a randomly generated Q-table for each new run. Instead of new Q-table each time, the past Q-table can be referred as initial table for second run onwards.

TABLE I
STEPS OF THE HuMAQ ALGORITHM FOR SINGLE AGENT

| **First run** |
| :--- |
| *I. Generate Q-table with random values* |
| *Repeat (forever)* |
| *{* |
| *1. Read the sensor data* |
| *2. Decide the state ($x_i$) of the robot* |
| *3. Select the action (a) according to the state and Q-values of the corresponding action* |
| *4. Execute the selected action* |
| *5. Read the sensor data* |
| *6. Compute the reward/ punishment & reinforce the state − action pairs(s)* |
| *7. Update Q-table as final table* |
| *}* |
| **Second run onwards** |
| *II. Initialization: Initial Q-table = Final Q-table* |
| *Repeat (forever)* |
| *{* |
| *1. Read the sensor data* |
| *2. Decide the state ($x_i$) of the robot* |
| *3. Select the action (a) according to the state and Q-values of the corresponding action* |
| *4. Execute the selected action* |
| *5. Read the sensor data* |
| *6. Compute the reward/ punishment & reinforce the state − action pairs(s)* |
| *7. Update Q-table as final table* |
| *}* |

In most cases, a single-agent is performing all the interactions between the system and environment necessary for Q-learning [refer fig. 1]. But for complex situations, (say a robot with large nos. of sensors) the single-agent may not perform well. Such complex cases can be addressed by multi-agent architecture. In such cases nos. of agents individually and independently works to achieve own goal, which is a sub-part of the overall goal. When all the agents individually achieve their own goals, the main goal is achieved. Here different reactive agents have been used and the Q-tables for different agents are generated using gradual learning, except the first run.

The initial concept of Q-learning was proposed using single-agent. But later on with the advancement of robotics, multi-agent system has been developed for complex situations using several sensors. Different agents are responsible for different sensors.

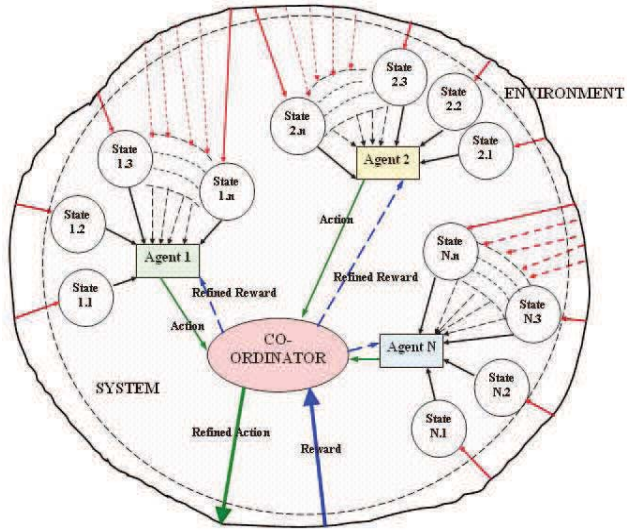

Fig. 2. Proposed Generalized Multi-agent Q-learning for a System. The system is divided in two main parts: different Agents and the Co-ordinator. System interacts with the environment with the help of Co-ordinator.

Human-like Gradual Multi Agent Q-learning is the proposed new approach of Q-learning for multi-agent systems. The human-like gradual learning has already been explained in the above paragraphs. The human-like gradual Q-learning for various multi-agent systems will be discussed in the following paragraphs. This is simpler than any other multi-agent Q-learning. As shown in fig. 2, the system interacts with the environment with the help of a 'Co-ordinator'. It is one the main parts of HuMAQ.

The system has 'N' numbers of agents responsible for different states generated by the sensors and by some internal states. For a simple system, each agent can take care of 2 states of the robot. For a more complex system, each agent can take care of n states. On the basis of the different states, each agent chooses different actions and sends to the Co-ordinator. This co-ordinator arranges the agents according to the priority as shown in fig. 3. For example if there are four agents; hunger, goal-seeking, obstacle-avoidance and line-following; they can be arranged as hunger, obstacle-avoidance, goal-seeking and line-following according to the priority. Hunger should have the highest priority, as no system can move or work without energy.

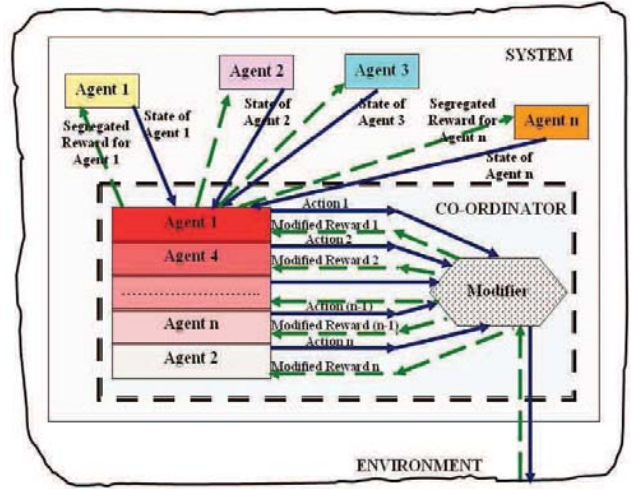

Fig. 3. The Proposed Generalized Co-ordinator for HuMAQ. The main part of the co-ordinator is the modifier. It modifies the actions for different agents depending upon the priority and distributes the reward accordingly

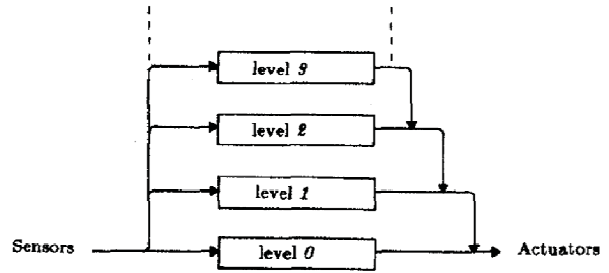

Fig. 4. Subsumption Architecture is a layered based control system for robots as proposed by Rodney Brooks. The top layers have higher priority over the bottom layers.

The next priority is of the obstacle-avoidance agent, as obstacle avoidance is the second most important issue after the power supply for real time explorations. The last two agents according to the priority are goal-seeking and line-following respectively, as goal-seeking is the main objective of any autonomous field exploration rather than the line-following. This concept of priority is based on Brook's Subsumption Architecture. As shown in fig. 4, this is a layer-based control for robots and distributed and parallel methods to connect sensors and actuators in robots. Each parallel layer (layer 2, layer 1 and layer 0) is made up of simple processors, called Augmented Finite State Machines (FSM). The most important aspects of FSMs are that outputs are simple function of inputs and local variables; inputs can be suppressed and outputs can be inhibitated.

Co-ordinator has an important sub-part known as modifier. The modifier refines the actions proposed by different agents depending upon their priorities. For example, if the goal-seeking agent has directed the robot to move in forward direction, but the obstacle-avoidance agent has detected obstacle in the front, the modifier searches the second point from where the goal is nearer and refines the action as move forward in left or move forward in right. The modifier tries to find out a common intersecting area (as shown in fig. 5) where the states from most of the different agents are available and then refines the action from the descending list of states from that common space accordingly. It is not mandatory that all the states from all the agents are present as in any instance of time all the agents may not be active. With the increase in number of agents, the complexity of the co-ordinator and modifier also increases.

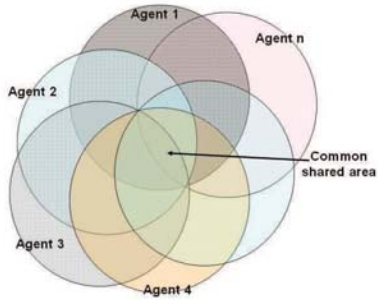

Fig. 5. The common shared area by all the agents is chosen by the modifier for consideration.

The refined action is performed by the actuators and the obtained reward is again sent to the modifier to distribute them as per the priority of the agents only to those, which were active for obtaining the particular reward. So, instead of a single reward as in the case of Q-learning, here a distributed and gradually reducing weightage system has been used. Such a variable weightage system has been used to accommodate the difference in priority of the agents. For example, if there are four agents and the total reward obtained is $R$ for any instance of time, when all the four agents were responsible for a refined action, it can be distributed as mentioned in eqn. 4 below.

$$R = \omega_1 R_1 + \omega_2 R_2 + \omega_3 R_3 + \omega_4 R_4 \qquad (4)$$

Where $\omega_1 > \omega_2 > \omega_3 > \omega_4$; $\omega_1 + \omega_2 + \omega_3 + \omega_4 = 1$; $\omega_1, \omega_2, \omega_3, \omega_4$ are the distributed weightage for rewards for agent 1 (highest priority), agent 2, agent 3 and agent 4 (low priority) respectively. $R_1, R_2, R_3$ and $R_4$ are the individual rewards for agent 1, agent 2, agent 3 and agent 4 respectively obtained directly due to system-environment interactions. With these separate rewards $\omega_i R_i (i \in 1,2,3,4)$, the separate Q-tables for different agents update themselves gradually. After a long run in any field exploration, if the system needs to be recharged or powered off for any reason, the final Q-table at that instance is stored and will be used as initial table for next run onwards. More details about this have been given in the following example.

### C. Simplified Methodology for Field Exploration

It has been noted that even if only a few sensors are used, the numbers of possible states become large. In an indoor exploration (goal seeking) with the help of light sensor, ultrasonic sensor, compass and battery-level detector (four different agents), nos. of states possible are in the order of $10^8$ or more. It may be very difficult to implement such a system in real time robots as it will need huge computational power and complexity. To simplify this model only three very necessary sensors have been used in real experiments. Also several assumptions have been made and as a result the total numbers of states have reduced to very minimum (in the order of $10^3$).

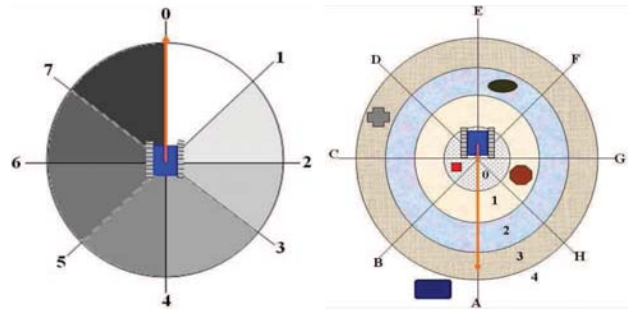

Fig. 6. (a) The numbering of the sectors is done in the clockwise direction with reference to the front direction of the robot. The numbers of the sectors are considered for denoting the brightness of the environment around the robot. (b) The surroundings of the robot can be differentiated as near, near-far, far, farther and farthest depending upon its distance. All these circles can again be divided into eight sectors. Obstacles with different colours and shapes have been found at different distances and different sectors.

Sensors are used to get the complete information of all the disturbances of the environment around a robotic system. In ideal case any sensor should have 360° views of the environment. Here the sensors have been mounted on a rotating head to take the readings around the robot (360° view). Instead of continuous reading a sector wise reading is preferred. Continuous reading will show gradual change in the reading of light; where as the sector wise reading will depict sudden change from sector to sector. Smaller division will lead to more numbers of sectors and consequently more numbers of sectors will show only a little difference in analogue values of the readings and the sensor will take longer time to read the data. But in some cases (in a dynamic environment), this delay may cause a problem. Similarly, for a few divisions, only a numbers of sectors with larger area will be generated and such large sectors will make a huge deviation in the trajectory of the robot and this may lead the robot to wrong direction and the time for error correction will make a cumulative effect on the mission time.

For a goal-directed field exploration, the minimum sensors to be used are (1) for monitoring the voltage of the battery (2) for detecting the goal (3) for avoiding the obstacles. In this case three different agents have been used for describing the states of these sensors as Hunger, Goal-seeking and Obstacle-avoidance respectively for a successful exploration.

*1) Agent – Hunger:* As already described, the agent hunger is responsible for the monitoring the battery-voltage level and denotes its sub-states as above the threshold voltage (can be denoted by '0') or below the threshold voltage ('1'). The probable actions for this agent shut-down the system and do not shut-down.

*2) Agent – Light-seeking:* Considering all the aspects of both smaller and larger sectors, here the (imaginary) circle around the robot is divided into 8 equal sectors each making an angle of 45° at the centre. The light sensor making reading only on boundary line (45° apart from each other) and differentiates the sectors according to the intensity of the light and the boundary line of the sectors are assigned values from 0 to 7 in the clockwise direction with reference to the front direction of the robot (shown with red coloured arrow line in fig. 6(a)). In actual case the arrangements of bright and dark sectors do not occur regularly, there may be repetition and irregular arrangements.

From the above convention, it is clear that the brightest sector can have 8 values (0 to 7). For example, say the brightest zone is the first sector i.e. with value 0. For this single value, the next brighter zone can have 8 values. So, for these two sectors together, for the single value of the first sector, $1 \times 8$ possible combinations can occur. Again the next bright sector can have 8 values. For these three sectors $1 \times 8 \times 8$ (= $8^2$=64) values are possible, for a single value of the first sector. For all the 8 values of the first sector, there can be $8 \times 8 \times 8$ (=$8^3$) combinations for the three sectors. For 8 sectors total $8^8$ numbers of combinations are possible. This refers to 16777216 numbers of possible states of the robot. But such a large combination of states will lead to computational complexity and require huge memory power and operational time.

TABLE II
State Table for the Goal-seeking Agent

| No. of the maximum intensity | No. of the second maximum | No. of the third maximum | Repetition |
|---|---|---|---|
| grid | grid | grid | |
| Percept ID 1 | Percept ID 2 | Percept ID 3 | Percept ID 4 |
| 0 | 1 | 2 | 0 |

To overcome the above difficulties, instead of considering all the 8 sectors, three sectors with consecutive maximum brightness values are considered. So, total $8^3$ ($8 \times 8 \times 8$) i.e. 512 numbers of states are possible instead of $8^8$. As the goal of the exploration is to search the brightest part of the environment, constant movement is essential for the robot. It may also happen that the robot is stationary and as a result the reading of a particular sector is constant for the last few couple of minutes. This is defined as another percept and it can be denoted as 'whether the system has five repeated readings or not'. If it is 'Yes', the value 1 is assigned and for 'No', 0 is assigned. So, in total 1024 ($512 \times 2$) numbers of states are possible. The above table (table II) shows that the state of the robot is denoted by a 4 digit number. Here 4 nos. of percepts are considered having ID 1, 2, 3, 4.

This entire phenomenon is taken care of by the goal-seeking agent (agent 1). The detail schematic diagram for only the goal-seeking agent is given in fig. 7.

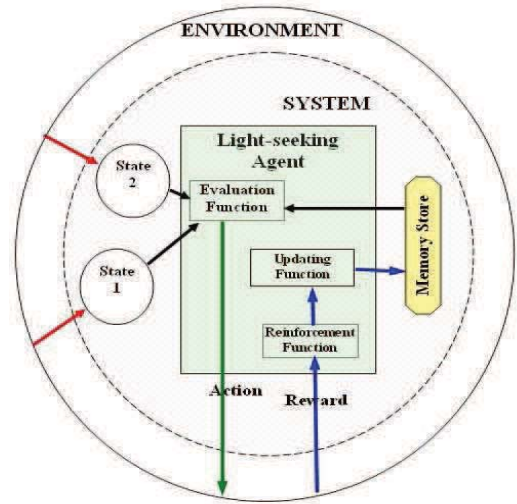

Fig. 7. The detail schematic diagram for the operation of the light-seeking agent (agent 1) shows that not only the system (robot) and the environment are considered, but also the memory of the robot is associated for determining the states.

The actions for this agent are given in table III. These actions are in terms of rotation of the robot along the sectors. As eight sectors have been considered for measuring the sensor values of the environment, there should also be eight rotations starting from 0° to 315° in steps of 45° as shown in table III. If the brightest zone is sector 0, (i.e. along the forward direction of the robot at current position) the robot will not rotate. But if the brightest zone is sector 7, the robot will rotate through an angle of 315° (i.e. 7 × 45°) clockwise. The goal - seeking agent chooses the action from the Q-table and sends to the co-ordinator.

TABLE III
ACTION TABLE FOR GOAL-SEEKING AGENT

| Actions | Action ID |
|---|---|
| No rotation | 0 |
| Clockwise rotation through **45°** | 1 |
| Clockwise rotation through **90°** | 2 |
| Clockwise rotation through **135°** | 3 |
| Clockwise rotation through **180°** | 4 |
| Clockwise rotation through **225°** | 5 |
| Clockwise rotation through **270°** | 6 |
| Clockwise rotation through **315°** | 7 |

1) *Agent – Obstacle-avoidance:* The ultrasonic sensor detects the obstacles and measures the distance. The obstacles only at the plane of the ultrasonic sensor are detected by the sensor. The surroundings of the robot can be referred as near, near-far, far, farther and farthest according to the distance and they can be denoted by circles (from 0 to 4) as shown in fig. 6(b). The approximate radii for these circles are given in table IV.

TABLE IV
DISTANCE TABLE FOR LOCATING OBSTACLES FOR OBSTACLE-AVOIDANCE
AGENT

| Relative position | Near | Near – far | Far | Farther | Farthest |
|---|---|---|---|---|---|
| Approximate distance (cm) | ≤ 50 | 50 - 100 | 100 - 150 | 150 - 200 | > 200 |
| Number of the zone | 0 | 1 | 2 | 3 | 4 |

Each individual circle surrounding the robot (circles around it) is again divided in 8 sectors (A to H). Total nos. of states for obstacles-avoiding agent can be found out in the similar manner as done in the case of goal-seeking agent. Suppose an obstacle is detected inside the first circle (no. 0) and in the second sector (no. B) as shown by a red square in fig. 6(b). So, for this single obstacle, there may be 5 possibilities (as there are only 5 circles) of an obstacle to be present in the next sectors i.e. first, second, third and fourth circle. This is to be noted that it is immaterial whether more objects are detected behind the first obstacle in the same sector. In such cases only the first obstacle (nearest to the centre of the circles) is considered. So together, 1 × 5 combinations are possible. Again, if the first sector is considered, total 1×5×5 (or $5^2$) combinations can be obtained. For all the sectors of the different circles, in total $5^8$ i.e. 390625 combinations (i.e. states) are available. But as in the earlier case, it is not possible to handle such large amount of data which require a lot of computational power, memory space and processing time.

For the simplicity, rather only those sectors of the circles are considered, where the brightness is maximum, second maximum and third maximum. Thus the total nos. of states reduces to $5^3$ i.e. 125 nos. instead of $5^8$ combinations. The robot should also move continuously to avoid trap. So, to avoid traps repeated readings of the robots should be avoided. This is denoted by another state: 'Repeated reading?'- Yes or No which is referred by 1 and 0. So, here 2 combinations are possible and in total 125 × 2 (=250) nos. states can be obtained. In this case the state table will also consist of 4 digits as given in table V. The state IDs along with their description is given in the table below.

The Avoid-obstacle agent locates the obstacles and finds out their position. This agent works independently with out any intervention by other agents. The action for this agent is given in table VI. Here all the three agents work independently and separately. But they are sharing information as the goal-seeking agent passes the information of three maximum brightness zones to the obstacle-avoidance agent.

Table V

The state Table for the Obstacle-avoidance Agent

| Obstacle in the Maximum Intense Zone | Obstacle in the Second Maximum Intense Zone | Obstacle in the Third Maximum Intense Zone | Repetition |
|---|---|---|---|
| Percept ID 1 | Percept ID 2 | Percept ID 3 | Percept ID 4 |
| 0 | 1 | 2 | 0 |

Table VI

Action Table for Obstacle-avoidance Agent

| Position of obstacles | Within 1st circle | Between 1st and 2nd circle | Between 2nd and 3rd circle | Between 3rd and 4th circle | Outside 4th circle |
|---|---|---|---|---|---|
| Speed (RPM) | 34 | 67 | 84 | 100 | 117 |
| Action ID | 0 | 1 | 2 | 3 | 4 |

*1) Co-ordinator and the Modifier:* All the sub-states and the recommended actions for different agents are forwarded to co-ordinator for selecting the refined action. The modifier chooses the refined actions against the above simplified states from the initial or updated Q-table based on the Q-values.

All the three agents are sending data (sub-states and action) simultaneously to the Co-ordinator. As the agent Hunger, is of highest priority, the co-ordinator sorts out this agent first and forward its recommended action and sub-state to the modifier. If the recommended action is 'shut-down system', the same is passed for action. No other command by other agents will be considered. But if the action is 'do not shut down', the modifier looks for the next priority agent. The next priority agent is the Goal-seeking agent. The agent gathers the data (brightness) from the surroundings and arranges only three of them (numbers of sectors with different brightness) in descending order starting with the maximum. Suppose the sub-states for such a case is '2 − 3 − 1'. The action related to the maximum brightness is rotation through an angle (= number of the sector × 45°) in the clockwise direction (as the reading has been taken in a step of 45° in the clockwise direction). Here it will be rotation through 90° (2 × 45°). The sub-states and the action are passed to modifier for further action. Obstacle-avoidance agent is the agent with last priority. This agent gathers the data about the presence of obstacles in the sectors and circles (1st, 2nd, 3rd and

4th) and compares them with the sectors of first, second and third maximum brightness. Then the agent passes the sub-states and action to the Co-ordinator for refinement. Here for 5 different zones (within 1st , 2nd , 3rd and 4th circle and outside 4th circle) and 3 different brightness regions, total 215 combinations are possible for presence ('1') or absence ('0') of obstacles. The modifier chooses the action according to following rules.

Rule - I: If there is an obstacle within the first circle of the brightest zone, go to second bright zone. If there is an obstacle also within the first circle of the second bright zone, go to the third bright zone. If there is an obstacle with in the first circle of the third bright zone, the system may stand still.

Rule - II: If the obstacle is not within the first circle, but within second or third or fourth circle of the considered bright zone, the system will use modified speed as given in table VI. In this regard a point may be noted that if there is an obstacle within the first circle of the brightest zone; another obstacle within the second circle (more specifically between first and second circle) of the second bright zone and no obstacle in the third bright zone, the modifier will make the system move along the second bright zone with modified speed.

The reward obtained due to the past action is updated in the Q-table for respective agents. Here two cases may happen.

Case-I: When the agent hunger detects that the voltage level is below the predefined threshold limit, the reward will be totally updated to the Q-table of the agent hunger, as this agent solely defines the action.

Case-II: When the agent hunger detects that the voltage level is above the predefined threshold limit, the reward is distributed with reducing weightage as mentioned in equation 2.

Here as only three agents are active, goal-seeking, obstacle-avoidance and direction-correction, the rewards are distributed with the weightages 0.6, 0.3 and 0.1 respectively. The reward for goal-seeking is calculated as:

$$[(L_{\max})_i - (L_{\max})_{i-1}] \qquad \text{.................} (3a)$$

Where $(L_{\max})_{i-1}$ = maximum light value of the previous state $(i-1)$ in the direction of motion of the robot and $(L_{\max})_i$ = maximum light value in the $(i)$ current state for that same direction. Similar policy is also adapted for obstacle-avoidance agent as mentioned in eqn. 3(b).

$$[(D_{\max})_i] \qquad \text{.................} (3b)$$

Where $(D_{\max})_i$ = maximum distance of the obstacle (if present or '255' if not present. 255 is the maximum range of the ultrasonic sensor) in the same direction in which the robot has moved to reach the $(i)$ current state.

For no repetition in any of the above mentioned agents, a positive scalar reward of magnitude '10' or '100' and for repetition a negative reward (i.e. punishment) of '-20' or '-200' have been used. The punishment should be always higher than the reward so as to stop selecting any wrong policy. The high value of punishment will completely abandon the associated action. These rewards multiplied with the weightage factors are updated in the respective Q-tables of different agents.

## V. EXPERIMENTAL RESULTS AND DISCUSSIONS

### A. Experiment

Different experiments in the field of Behavior-based robotics, Reinforcement learning has been carried out in different environments using different robotic systems. Initially it was a challenge to identify the right Behavior-based architecture for implementation. For this purpose two most popular architectures (Subsumption architecture and Motor-schema theory) have been tested in simulated environment. Using the identified architecture, the next series of experiments have been carried out in simulated environment (fig 8(a)), indoor environment (fig 8(b)) and different outdoor terrains (fig 8(c)-(f)). Most of these experiments include only three agents: hunger, goal-seeking and the obstacle-avoidance using the battery-level detector, light sensor and ultrasonic sensor respectively. The data of the experiments have been saved on the robot itself and later downloaded for further processing.

The experiments with HuMAQ are carried out in five trial runs on different outdoor terrains each with five sets. The first set of each trial run uses a randomly initialized Q-table and from second run onwards, final updated Q-table of the previous set is used as the initial Q-table of the next set.

### B. Experimental Systems

In these sets of experiments four different sets of robots have been used. Initially the experiments have been carried out with indigenously developed three different robots: ARBIB (Autonomous Robot Based on Intelligent Behaviors) – I, II and III. The testing for HuMAQ has been mostly carried out on LEGO Mindstorm NXT due to its simple and perfect analog sensors. The above figures [9(a) – (d)] show the different robots used during various experiments.

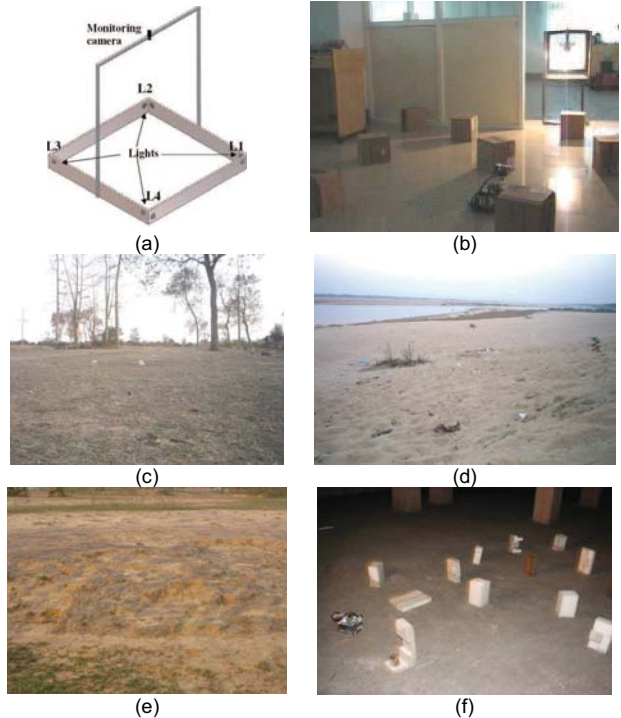

Fig. 8. Environments where different experiments have been carried out (a) Simulated environment (b) Indoor environment (c) Outdoor terrain – Plain grassland (d) Outdoor terrain – Loose sand (e) Outdoor terrain – Laterite tableland (f) Outdoor terrain- Roof of the laboratory during night

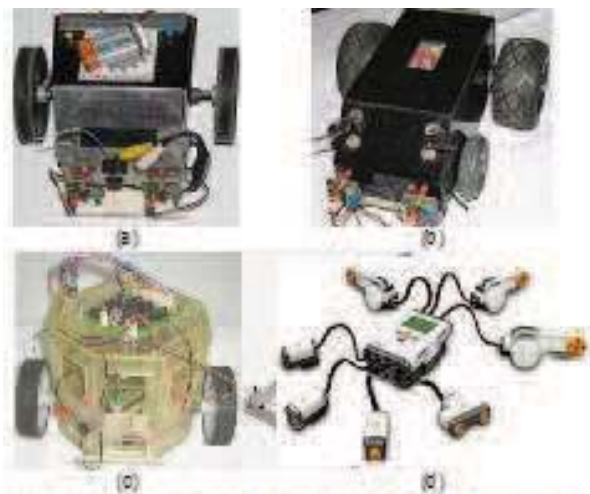

Fig. 9. Different Robotic Systems used in the Experiments (a) ARBIB-I (b) ARBIB-II (c) ARBIB-III (d) LEGO Mindstorm [courtesy LEGO Group]

| Items | Set A | Set B | Set C | Set D | Set E |
|---|---|---|---|---|---|
| Value Increased | 42 | 60 | 71 | 78 | 82 |
| Value Decreased | 19 | 19 | 19 | 19 | 19 |
| Total Change | *61* | *79* | *90* | *97* | *101* |

### C. Results and Discussions

After the successful completion of the exploration in indoor environment using HuMAQ, the data are downloaded for analysis. The first set (Set A) of the trial run is done with a randomly generated Q-table. In this set, 54 nos. of updates are performed to reach the goal. The second set also took 54 nos. of updates to reach its goal, but started with the final table of set A as initial table. The nos. of updates for other sets of the same trial run and the approximate time to reach the goal has been given in table VII. When these numbers of updates (or the approximated time) are plotted, produces a graph as shown in fig. 10(a) which is gradually decaying down. This gradual decrease in nos. of updates or approximated time indicates that learning is occurring in real time. If the experiment is carried out for few more sets, the curve will become flat. It means that the learning has reaches the optimality.

Table VIII and IX show the cumulative change of Q-values in the Q-table with respect to its initial values for goal-seeking (using light sensor) and obstacle-avoidance (using ultrasonic sensor) agents respectively. The different between two consecutive columns (i.e. sets) gives the particular change of the Q-values occurred in the later. If these particular values for different columns (sets) are compared, it will also show a decreasing tendency and this proves the efficiency of the learning.

Table VII

Numbers of Updates and Approximate Time Required (Sec) for Reaching the Goal in an Indoor Environment using HuMAQ

| | Set A | Set B | Set C | Set D | Set E |
|---|---|---|---|---|---|
| Nos. of Updates | 57 | 56 | 48 | 38 | 34 |
| Time Required (Sec) | 1141 | 1122 | 964 | 762 | 685 |

Table VIII

Cumulative Change of Q-values (compared to intial value) in the Q-table for the Goal-seeking Agent (Light sensor) in Indoor Environment

Table IX

Cumulative Change of Q-values (compared to intial value) in the Q-table for the Obstacle-avoidance Agent (Ultrasonic sensor) in Indoor Environment

| Items | Set A | Set B | Set C | Set D | Set E |
|---|---|---|---|---|---|
| Value Increased | 16 | 31 | 36 | 41 | 44 |
| Value Decreased | 9 | 9 | 9 | 9 | 9 |
| *Total Change* | *25* | *40* | *45* | *50* | *53* |

Table X

Durations (in sec) for reaching the goal

| Types of Terrains | Set A | Set B | Set C | Set D | Set E |
|---|---|---|---|---|---|
| Plain Grass land | 1382 | 1203 | 784 | 904 | 781 |
| Sandy river bank | 1323 | 1081 | 1022 | 904 | 900 |
| Hard concrete floor | 1201 | 1025 | 845 | 802 | 761 |
| Laterite table land | 902 | 783 | 722 | 601 | 600 |
| Hard concrete floor at night | 1261 | 1203 | 904 | 843 | 842 |

Table XI

Cumulative Change of Q-values (compared to intial value) of the various state-action pairs for the Goal-seeking Agent (Light sensor)

| Types of Terrains | Items | Set A | Set B | Set C | Set D | Set E |
|---|---|---|---|---|---|---|
| Plain Grass land | *Value Increased* | 46 | 56 | 62 | 65 | 66 |
| | *Value Decreased* | 19 | 19 | 19 | 19 | 19 |
| | *Total Change* | *65* | *75* | *81* | *84* | *85* |
| Sandy river bank | *Value Increased* | 35 | 46 | 50 | 52 | 54 |
| | *Value Decreased* | 20 | 20 | 20 | 20 | 20 |
| | *Total Change* | *55* | *66* | *70* | *72* | *74* |
| Hard concrete floor | *Value Increased* | 43 | 52 | 60 | 70 | 79 |
| | *Value Decreased* | 19 | 19 | 19 | 19 | 19 |
| | *Total Change* | *62* | *71* | *79* | *89* | *98* |
| Laterite table land | *Value Increased* | 30 | 36 | 42 | 47 | 48 |
| | *Value Decreased* | 26 | 26 | 26 | 26 | 26 |
| | *Total Change* | *56* | *62* | *68* | *73* | *74* |
| Hard concrete floor at night | *Value Increased* | 47 | 57 | 59 | 64 | 66 |
| | *Value Decreased* | 17 | 17 | 17 | 17 | 17 |
| | *Total* | *64* | *74* | *76* | *81* | *83* |

| Types of Terrains | Items | Set A | Set B | Set C | Set D | Set E |
|---|---|---|---|---|---|---|
| | *Change* | | | | | |

Table XII

Cumulative Change of Q-values (compared to intial value) of the various state-action pairs for the Obstacle-avoidance Agent (Ultrasonic sensor)

| Types of Terrains | Items | Set A | Set B | Set C | Set D | Set E |
|---|---|---|---|---|---|---|
| Plain Grass land | *Value Increased* | 18 | 26 | 30 | 31 | 33 |
| | *Value Decreased* | 6 | 6 | 6 | 6 | 6 |
| | *Total Change* | *24* | *32* | *36* | *37* | *39* |
| Sandy river bank | *Value Increased* | 15 | 24 | 28 | 31 | 33 |
| | *Value Decreased* | 5 | 5 | 5 | 5 | 5 |
| | *Total Change* | *20* | *29* | *33* | *36* | *38* |
| Hard concrete floor | *Value Increased* | 13 | 21 | 25 | 26 | 28 |
| | *Value Decreased* | 2 | 2 | 2 | 2 | 2 |
| | *Total Change* | *15* | *23* | *27* | *28* | *30* |
| Laterite table land | *Value Increased* | 11 | 13 | 15 | 19 | 22 |
| | *Value Decreased* | 5 | 5 | 5 | 5 | 5 |
| | *Total Change* | *16* | *18* | *20* | *24* | *27* |
| Hard concrete floor at night | *Value Increased* | 12 | 17 | 19 | 21 | 22 |
| | *Value Decreased* | 9 | 9 | 9 | 9 | 9 |
| | *Total Change* | *21* | *26* | *28* | *30* | *31* |

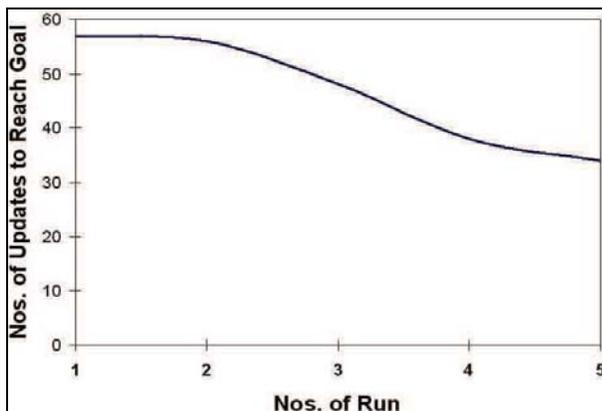

Fig. 10(a). The nos. of updates to reach the goal Vs. nos. of trial runs shows a gradually reducing nature which in fact supports the actual learning using HuMAQ in indoor environment.

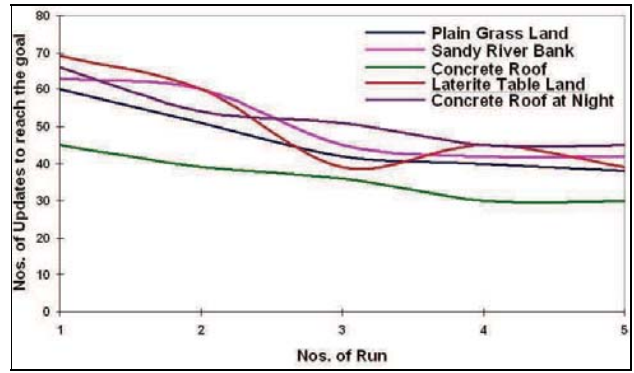

Fig. 10(b). The time to reach the goal Vs. run nos. shows a good acceptability of HuMAQ for autonomous exploration in outdoor terrains. As expected, the initial time was higher than the later.

In similar manner, the different data recorded during the outdoor field explorations and the extracted data/ information generated thereafter have been given in the above tables/ sections. The durations for reaching the goals (calculated thereafter) different sets of the different trial runs have been given in table X.

The cumulative change of Q-values in the Q-table with reference to the initial values for both goal-seeking and obstacle-avoidance agents have been given in table XI and XII respectively.

One important issue for any learning algorithm/ theory is that if learning is applied in the same field for same objectives repeatedly, the learning time/objective fulfilment time should reduce. This point has also been fulfilled in the case of application of HuMAQ for autonomous explorations. It is clear from table X that the time for reaching the goal from the same starting point, are gradually reducing in nature. For example, if the experiment over the plain grass land is considered, the initial nos. of updates to reach the goal is 66 with randomly generated Q-table and the final (set E) nos. of updates to reach the goal is 45 using the concept of gradual learning. For all other intermediate sets, the update-counts are lying between 66 and 45.

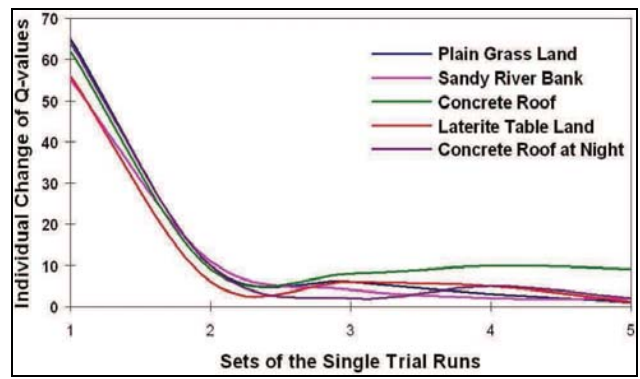

Fig. 11. The individual change of Q-values of the goal – seeking agent (light sensor) for different sets of the different trial runs proves the convergence of the learning.

### D. Performance of the Learning Algorithm

As described in [25], 'optimality is usually an asymptotic result and so convergence speed is an ill-defined measure'. As optimality may not be well defined, more practical measure is speed of convergence to near – optimality. This indicates that someone should define how near the optimality is sufficient. The speed of convergence or the rate of convergence can be defined [31] as follows. If a sequence of numbers ($a_1, a_2, a_3,$ ……….. $a_k$ ) is considered and this sequence converges to the number L, it can be said that this sequence converges linearly to L, if there exists a number, $\mu$ =(0, 1) such that

$$\lim_{k \to \infty} \frac{|a_{k+1} - L|}{|a_k - L|} = \mu$$

This $\mu$ is known as the speed or rate of convergence. If $\mu$ =0, it is said the sequence converges linearly and if $\mu$ =1, the sequence converges sub-linearly.

If the updates of the trial runs for explorations in outdoor environment are considered as sequence of numbers, for example, for the first trial run (experiment on plain grass land) the updates are 69 (set A), 60 (set B), 39 (set C), 45 (set D), 39 (set E) and if the lowest of all the updates of the five sets of a trial run is considered as optimal one or near optimality, the speed or rate of convergence is calculated as shown below in table XIII. In the above mentioned first trial run the optimum value is 39 (i.e. L) and all the calculations in absolute values are done with respect to this value. Different optimal values for different fields are set as the experiments have been carried out in different environmental conditions.

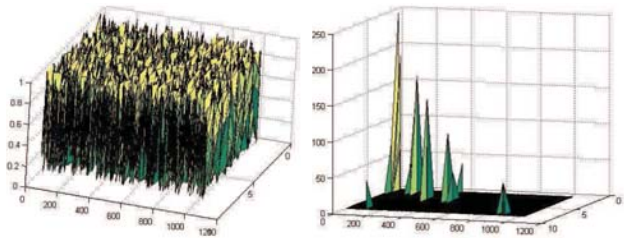

Fig. 12(a). Initial Q-table for light sensor during testing on plain grass land (b) Final Q-table for light sensor during testing on plain grass land

Table XIII

Speed of Convergence for the Updates for Outdoor Exploration

| Types of Terrains | Set A | Set B | Set C | Set D | Set E |
|---|---|---|---|---|---|
| Plain grass land | 0.7 | 0 | - | 0 | - |
| Sandy river bank | 0.4 | 0.7 | 0 | - | - |
| Hard concrete floor | 0.6 | 0.3 | 0.5 | 0 | - |
| Laterite table land | 0.6 | 0.7 | 0 | - | - |
| Hard concrete floor at night | 0.9 | 0.2 | 0 | - | - |

The speed of convergence lies between 0 and 1 that means this is linear in nature. The cell is left blank where the value is undefined. The higher values of speed of convergence refer to fast reaching of the optimality and the lower values lead to slow reaching. However a medium speed of convergence is preferred as both higher and lower values may be misleading [25].

### E. Comparison with Other Reinforcement Learning Methods

Dynamic programming and Monte Carlo Methods are the two well known methods under reinforcement learning. Dynamic programming refers to a collection of algorithms that can be used to compute optimal policies given a perfect model of the environment as a Markov decision process [1]. Monte Carlo methods are ways of solving the reinforcement learning problem based on averaging sample returns [1]. Experiments have been carried out using Dynamic Programming and Monte Carlo methods in the indoor environment with same environmental and experimental parameters. The results obtained are plotted in fig. 13 (a) and (b). Both the graphs clearly show (even if only 4 iterations are considered for comparison in all cases) that they do not follow similar pattern as in the case of HuMAQ and thus could not provide efficient learning. In both cases the initial learning time is less than the final learning time. Human-like gradual learning provides better result than these methods.

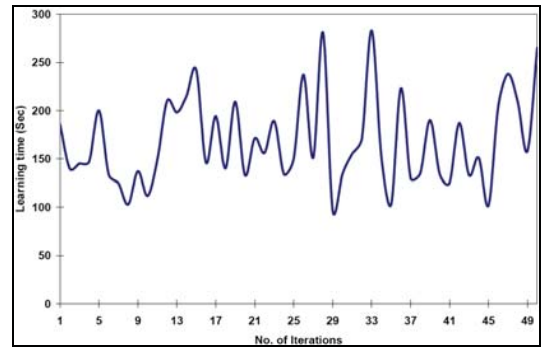

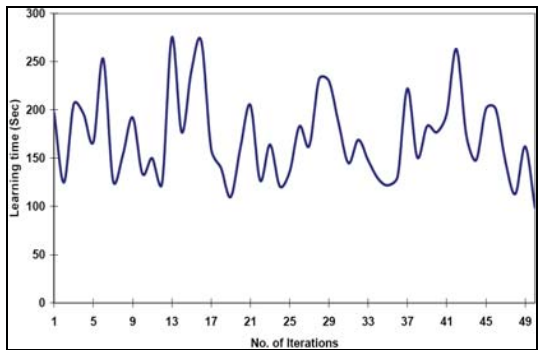

Fig. 13. Comparison with other Reinforcement Learning Methods: Curves for learning time vs. no. of iterations to reach the goal using (a) On-policy Monte Carlo Method [Top] (b) Dynamic Programming

## VI. CONCLUSION

The behavior-based robotics has opened up a new field of robotics which uses the Sense →Act paradigm for achieving the low-level intelligence easily. Four different behavior-based architectures are popular, out of which only two (Subsumption architecture and the Motor-schema Theory) have been tested to implement here.

The reinforcement learning is a suitable machine learning approach for implementing in mobile robots and this is a direct system-environment interaction based on reward/ punishment policy. Q-learning a sub-issue of reinforcement learning uses delayed reward/punishment for the previous action. It uses a state-action mapping table based on the different conditions of the sensors (states) and commands to the actuators (action). Agent is a software entity which performs an assigned task independently. In most cases Q-learning uses single agent to perform all the tasks for learning. Use of multiagent is preferred to handle with large numbers of sensors and therefore large numbers of state-action mapping.

As revealed from various literatures, human learning is cumulative and gradual in nature and as time passes the learning time for the same topic/ subject reduces. This concept of gradual learning can be incorporated with the multiagent Q-learning to get a better performance in outdoor terrain explorations by mobile robots.

Here a new approach of multiagent reinforcement learning has been proposed using the concept of Subsumption architecture, a well-known behavior-based architecture, and the gradual learning technique for autonomous exploration. It uses different agent on single system, not different agents on different systems. The testing of HuMAQ in different terrains, different static and partial dynamic conditions reveal the acceptance of the algorithm for autonomous explorations by mobile robots. Also the measurement of the performance of the learning algorithm has been done from the proof of convergence.

### ACKNOWLEDGMENT

This work has been carried out as a part of SIP-24, an 11[th] Five Year Project funded by CSIR, Govt. of India.

Authors also thank CMERI personals who have directly and indirectly helped to complete this project. It will be injustice if the constant help and effort made by Mr. Amit Mandal, research scholar in this project, and Mr. A. Mahapatra, scientist, CMERI is not acknowledged.